\documentclass[sigconf,natbib=true,anonymous=false]{acmart}
\AtBeginDocument{%
  \providecommand\BibTeX{{%
    \normalfont B\kern-0.5em{\scshape i\kern-0.25em b}\kern-0.8em\TeX}}}

\copyrightyear{2024}
\acmYear{2024}
\setcopyright{acmlicensed}\acmConference[SIGIR '24]{Proceedings of the 47th International ACM SIGIR Conference on Research and Development in Information Retrieval}{July 14--18, 2024}{Washington, DC, USA}
\acmBooktitle{Proceedings of the 47th International ACM SIGIR Conference on Research and Development in Information Retrieval (SIGIR '24), July 14--18, 2024, Washington, DC, USA}
\acmDOI{10.1145/3626772.3657745}
\acmISBN{979-8-4007-0431-4/24/07}

\usepackage{newfloat}
\usepackage{listings}
\usepackage{algorithm}
\usepackage{algorithmic}
\usepackage{adjustbox}
\usepackage{multirow}
\usepackage{booktabs}
\usepackage{tabularx}
\usepackage{color}
\usepackage{bm}
\usepackage{amsmath}
\usepackage{tikz}
\usepackage{xcolor}
\usepackage{tcolorbox}
\usepackage{microtype}

\begin{document}

\title{\texttt{I3}: \underline{I}ntent-\underline{I}ntrospective Retrieval Conditioned on \underline{I}nstructions}

\author{Kaihang Pan$^{\ast}$}
\affiliation{%
  \institution{Zhejiang University}
  \city{Hangzhou}
  \country{China}}
\email{kaihangpan@zju.edu.cn}

\author{Juncheng Li$^{\dagger}$}
\affiliation{%
  \institution{Zhejiang University}
  \city{Hangzhou}
  \country{China}}
\email{junchengli@zju.edu.cn}

\author{Wenjie Wang}
\affiliation{%
  \institution{National University of Singapore}
  \city{Singapore}
  \country{Singapore}}
\email{wenjiewang96@gmail.com}

\author{Hao Fei}
\affiliation{%
  \institution{National University of Singapore}
  \city{Singapore}
  \country{Singapore}}
\email{haofei37@nus.edu.sg}

\author{Hongye Song}
\affiliation{%
  \institution{DAMO Academy, Alibaba Group}
  \city{Hangzhou}
  \country{China}}
\email{hongye.shy@alibaba-inc.com}

\author{Wei Ji}
\affiliation{%
  \institution{National University of Singapore}
  \city{Singapore}
  \country{Singapore}}
\email{jiwei@nus.edu.sg}

\author{Jun Lin}
\affiliation{%
  \institution{ DAMO Academy, Alibaba Group}
  \city{Hangzhou}
  \country{China}}
\email{linjun.lj@alibaba-inc.com}

\author{Xiaozhong Liu}
\affiliation{%
  \institution{Worcester Polytechnic Institute}
  \city{Worcester}
  \country{United States}}
\email{xliu14@wpi.edu}

\author{Tat-Seng Chua}
\affiliation{%
  \institution{National University of Singapore}
  \city{Singapore}
  \country{Singapore}}
\email{dcscts@nus.edu.sg}

\author{Siliang Tang}
\affiliation{%
  \institution{Zhejiang University}
  \city{Hangzhou}
  \country{China}}
\email{siliang@zju.edu.cn}

\renewcommand{\shortauthors}{Kaihang Pan and Juncheng Li, et al.}

\begin{abstract}
Recent studies indicate that dense retrieval models struggle to perform well on a wide variety of retrieval tasks that lack dedicated training data, as different retrieval tasks often entail distinct search intents.
To address this challenge, in this work we leverage instructions to flexibly describe retrieval intents and introduce \texttt{I3}, a unified retrieval system that performs \textbf{\underline{I}}ntent-\textbf{\underline{I}}ntrospective retrieval across various tasks, conditioned on \textbf{\underline{I}}nstructions without any task-specific training. 
\texttt{I3} innovatively incorporates a pluggable introspector in a parameter-isolated manner to comprehend specific retrieval intents by jointly reasoning over the input query and instruction, and seamlessly integrates the introspected intent into the original retrieval model for intent-aware retrieval.
Furthermore, we propose \textbf{progressively-pruned intent learning}. It utilizes extensive LLM-generated data to train \texttt{I3} phase-by-phase, embodying two key designs: progressive structure pruning and drawback extrapolation-based data refinement.
Extensive experiments show that in the BEIR benchmark, \texttt{I3} significantly outperforms baseline methods designed with task-specific retrievers, achieving state-of-the-art zero-shot performance without any task-specific tuning. 

\end{abstract}

\ccsdesc[500]{Information systems~Retrieval models and ranking}

\keywords{Intent-introspective retrieval, progressive structure pruning, drawback extrapolation-based data refinement}



\maketitle
\section{Introduction}

\renewcommand{\thefootnote}{\fnsymbol{footnote}}
\footnotetext[1]{\ \ Work done when interning at Alibaba DAMO Academy.}
\footnotetext[2]{\ \ Corresponding Author.}
\renewcommand{\thefootnote}{\arabic{footnote}}

Information Retrieval~(IR) is a fundamental task with widespread applications not only in real-world scenarios such as web search~\cite{websearch} and digital libraries~\citep{dglibrary}, but also extending its significance to retrieval-augmented large language models (LLMs)~\citep{realm}. Recent dense retrieval models have demonstrated remarkable performance based on the transformer architecture in a manner of dual encoders.
Through the dual-encoders, they excel at encoding queries and documents into a shared representation space to facilitate semantic matching after the training with abundant annotated data~\citep{msmarco,nq}.

\begin{figure}[t]
    \includegraphics[width=\linewidth]{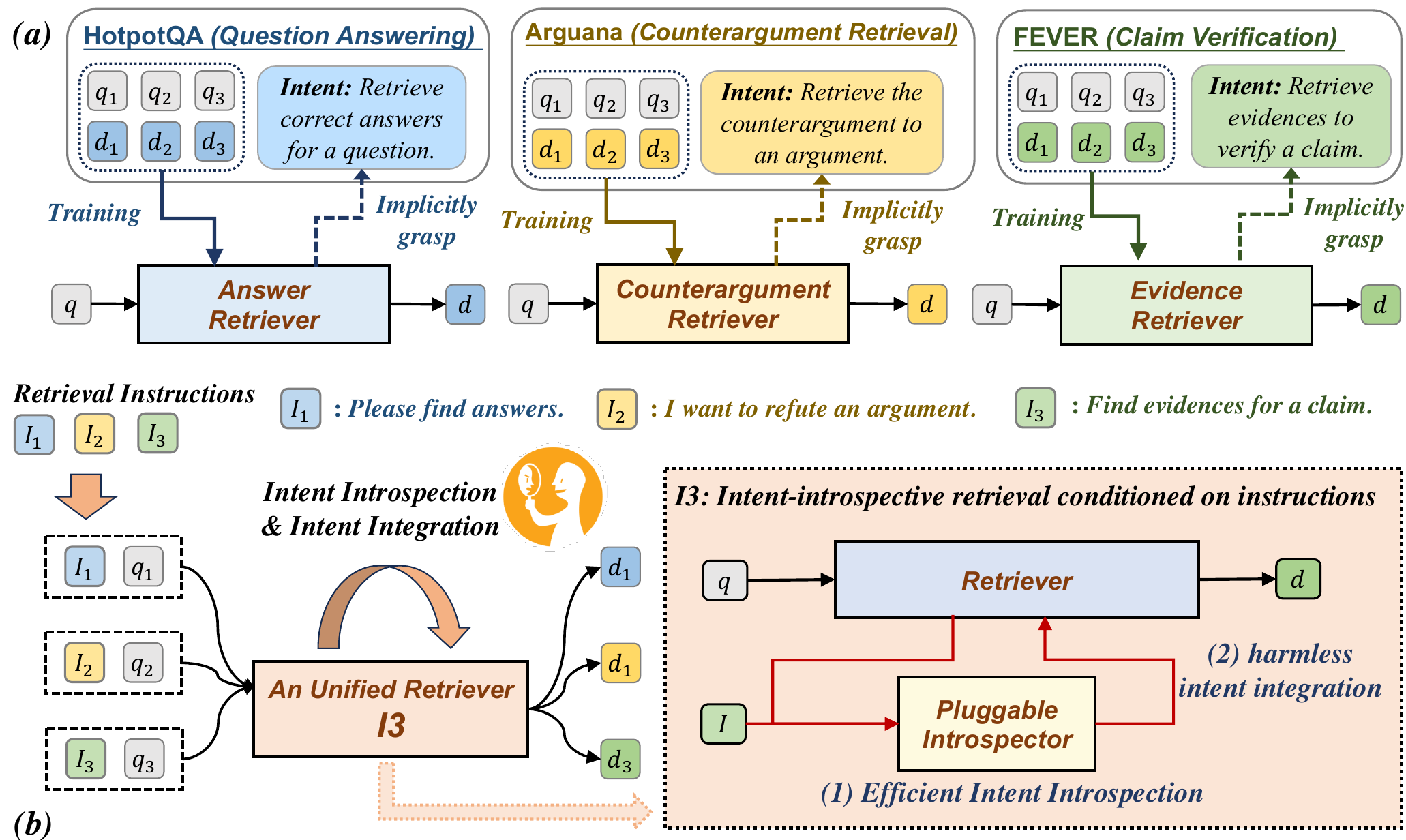}
    \vspace{-1.5em}
    \caption{(a) Existing methods require training a distinct retrieval model for each task to implicitly grasp specific retrieval intent. (b) \texttt{I3} directly handles different tasks through intent-introspective retrieval following user instructions.}\label{fig:intro}
\vspace{-1em}
\end{figure}

Nonetheless, this approach overlooks a critical fact that 
different retrieval tasks often entail varied search intents.
Recent studies~\citep{beir} indicate that existing dense retrieval models struggle to perform well on a wide variety of retrieval tasks that lack dedicated training data.
When encountering a novel retrieval task, sufficient annotated data is necessary for training retrieval models to implicitly grasp the search intent, as demonstrated in Figure~\ref{fig:intro}.a.  
Given the challenge of obtaining such annotated data, a recent work, Promptagator~\citep{promptagator} instructs LLMs to generate task-specific training data by presenting them with sets of 8 examples. 
Then, it utilizes the generated pseudo training data to train task-specific retrieval models for each distinct task.
While obtaining promising improvements, Promptagator necessitates the training of a distinct model to implicitly grasp the retrieval intent of each task, which limits the flexibility to seamlessly transfer across diverse retrieval tasks.

Instead of training a specific model for each retrieval task, it is desirable to enable pre-trained retrieval models to directly perform different tasks guided by instructions that flexibly describe retrieval intents in natural language~\cite{tart}.
To achieve this, 
an ideal approach can be divided into two steps:
(1) \textbf{\textit{effective intent introspection}}: deeply comprehending specific retrieval intent by jointly reasoning over the input query and instruction; (2) \textbf{\textit{harmless intent integration}}: integrating the introspected intent into the pre-trained retrieval model, while avoiding disrupting the model's original capabilities.
Based on this insight, we aim to achieve such flexible intent-aware retrieval from both model and learning perspectives.

\textbf{From the model perspective}, we introduce $\bf{I3}$, a generic approach that enables pre-trained retrieval models to effectively perform \textbf{intent-introspective retrieval} across various tasks, while simultaneously maintaining the pre-trained capabilities and avoiding any task-specific training.
Specifically,  given an arbitrary dense retrieval model with dual-encoders, \texttt{I3} retains the original encoders and innovatively develops a pluggable introspector in a parameter-isolated manner to preserve the inherent capability of retrieval models. 
The pluggable introspector subtly interprets the specific retrieval intent by jointly comprehending the given query and instruction. And the introspected intent is then seamlessly integrated into the original query encoder, 
which is harmless due to the parameter-isolated framework,
endowing retrieval models with a new facet of intent-aware retrieval conditioned on instructions.

\textbf{From the learning perspective}, to efficiently perform intent-introspective retrieval across a wide range of retrieval tasks, the pluggable introspector should (1) remain lightweight to avoid significantly affecting the time efficiency of retrieval, (2) effectively understand and perceive various retrieval intents within instructions. 
Therefore, we further propose \textbf{progressively-pruned intent learning} to iteratively train \texttt{I3} phase-by-phase, incorporating two key designs: 
(1) \textit{progressive structure pruning}, and (2) \textit{drawback extrapolation-based data refinement}.
Specifically, we harness the advancement of LLMs to automatically generate extensive data with instructions as the seed training dataset and then divide the total training into several phases. 
After each training phase, we prune the pluggable introspector into a sheared module, and simultaneously extrapolate the drawback of the current model to synthesize additional training data for refinement.
This allows \texttt{I3} to progressively comprehend a wide range of retrieval intents with an increasingly streamlined structure. 

Benefiting from the advancements in both model and learning perspectives, \texttt{I3} finally evolves into a unified and lightweight retrieval system, capable of efficiently following instructions to directly perform various retrieval tasks with diverse intents. 
Remarkably, in the BEIR benchmark \citep{beir}, feeding with only natural language descriptions of specific intent for each task, \texttt{I3} significantly outperforms competitive baselines without any task-specific fine-tuning. 
Overall, our main contributions are three-fold:

\begin{itemize}
	\item We innovatively propose \texttt{I3}, a generic and efficient approach that endows retrieval models with a new facet of intent-introspective retrieval following instructions, enabling them to directly perform diverse tasks without specific tuning.
	\item We devise progressively-pruned intent learning, which incorporates progressive structure pruning and drawback extrapolation based data refinement, training \texttt{I3} phase-by-phase with extensive LLM-generated instruction data.
    \item Experiments show that \texttt{I3} achieves state-of-the-art performance on BEIR benchmark under both zero-shot retrieval and reranking scenarios, without any task-specific tuning.
\end{itemize}

\section{Related Work}

\paragraph{Zero-shot Dense Retrieval. } Dense retrieval~\citep{dpr, latent, colbert, 10.1145/3626092, 10.1145/3511808.3557388} is widely adopted in information retrieval that demonstrates strong advantages over sparse retrieval methods~\citep{bm25}. However, \citet{beir} has shown that dense retrieval models still struggle to generalize to out-of-domain data and do not perform well in zero-shot settings, where no task-specific signals are available. To improve zero-shot dense retrieval, some existing works~\citep{gpl, qgen} bring in domain adaptation techniques and utilize documents from the target domain to generate corresponding pseudo queries for task-specific training. Other efforts~\citep{laprador, contriever, cocodr} focus on designing more efficient unsupervised contrastive learning paradigm, while some studies~\citep{gtr, cpt} attempt to enhance generalizability by scaling up retrieval models, among other approaches. 

Nevertheless, expecting retrieval models to perform well solely based on the query in a zero-shot setup can be inherently challenging as different tasks entail different retrieval intents. To reveal the retrieval intent, for each task Promptgator~\citep{promptagator} samples 8 query-document pairs and then generates pseudo retrieval data for training task-specific retrievers. In comparison, our method proposes a single unified retrieval model to perform various tasks following instructions without any task-specific tuning. 

Another recent work TART~\citep{tart}, leverages instructions to flexibly describe the retrieval intents. 
It collects 40 existing retrieval datasets and manually annotates them with instructions. 
Then TART directly inputs instructions into the query encoder, simply concatenating them ahead of the queries.
On this basis, it trains a multi-task retrieval model to perform retrieval tasks following instructions. 
However, TART faces the following challenges:
(1) the human-annotated training datasets are costly and suffer from limited diversity, with many of them sharing common retrieval intent~(\textit{e.g.}, retrieve correct answer to a question). 
(2) Given that retrieval models typically lack advanced in-context understanding capabilities, TART does not assure that the query encoder effectively understands the
retrieval intents within the instruction. And the addition of extra instructions also disrupts the original input format of the query encoder, potentially diminishing its inherent abilities.
In contrast, $I_3$ leverages LLMs to automatically generate instruction data comprising a wide range of retrieval intents with iterative drawback extrapolation-based refinement. 
It also preserves the inherent capability of retrieval models and efficiently empowers them with a new facet of intent-aware retrieval conditioned on task-specific instructions through a parameter-isolated architecture.

\section{Methodology}
In this section, we first give the preliminaries of dense retrieval (\S \ref{sec: 3.1}). 
Then, we introduce  \texttt{I3}, which performs intent-introspective retrieval with a parameter-isolated architecture (\S \ref{sec: 3.2}).
Finally, we elaborate on the progressively-pruned intent learning (\S \ref{sec: 3.3}). It capitalizes the advancement of LLMs to automatically generate extensive data (\S \ref{sec: 3.3.1}). And the LLM-generated data is then leveraged to train \texttt{I3} phase-by-phase, which incorporates two key designs: progressive structure pruning (\S \ref{sec: 3.3.2}) and drawback extrapolation-based data refinement (\S \ref{sec: 3.3.3}).

\subsection{Preliminaries}
\label{sec: 3.1}
Dense retrieval leverages a dual-tower architecture, consisting of a query encoder $E_Q$ and a document encoder $E_D$, to encode query $q$ and document $d$ into dense vectors. After obtaining the representations of both query and document, a similarity function (\textit{e.g.}, dot product) $s(q, d)$ is leveraged to calculate the relevance score between them:
\begin{equation}
\small
\begin{aligned}
    s(q,d) = \left \langle E_Q(q;\Theta_q), E_D(d;\Theta_d) \right\rangle
\end{aligned}
\end{equation}
To train the dual encoders, given a query and the relevant (positive) document $d^+$, a common approach is to sample irrelevant (negative) documents $d^-$, which can be either in-batch negatives, BM25 negatives~\citep{dpr}, or hard negatives mined by dense retrieval models~\citep{ance}. 
The objective is to maximize the probability of selecting the positive document over other negative documents via contrastive learning~\cite{simclr,ijcai2022-527}:
\begin{equation}
\small
\begin{aligned}
    p(d^+ | q,d^-) = \frac{e^{s(q,d^+)}}{e^{s(q,d^+)} + \sum e^{s(q,d^-)}}
\end{aligned}
\end{equation}
Furthermore, to realize retrieval with instructions that describe the retrieval intents in natural language, the instruction  $\mathcal{I}$ is incorporated into the query encoding. Then the query encoder can be formulated as $E^\prime_Q(q, \mathcal{I}; \Theta^\prime_q)$ with the similarity function defined as:
\begin{equation}
\small
\begin{aligned}
    s(q,\mathcal{I},d) = \left \langle E^\prime_Q(q, \mathcal{I}; \Theta^\prime_q), E_D(d;\Theta_d) \right\rangle
\end{aligned}
\end{equation}

\begin{figure}[t]
    \centering
    \includegraphics[width=\linewidth]{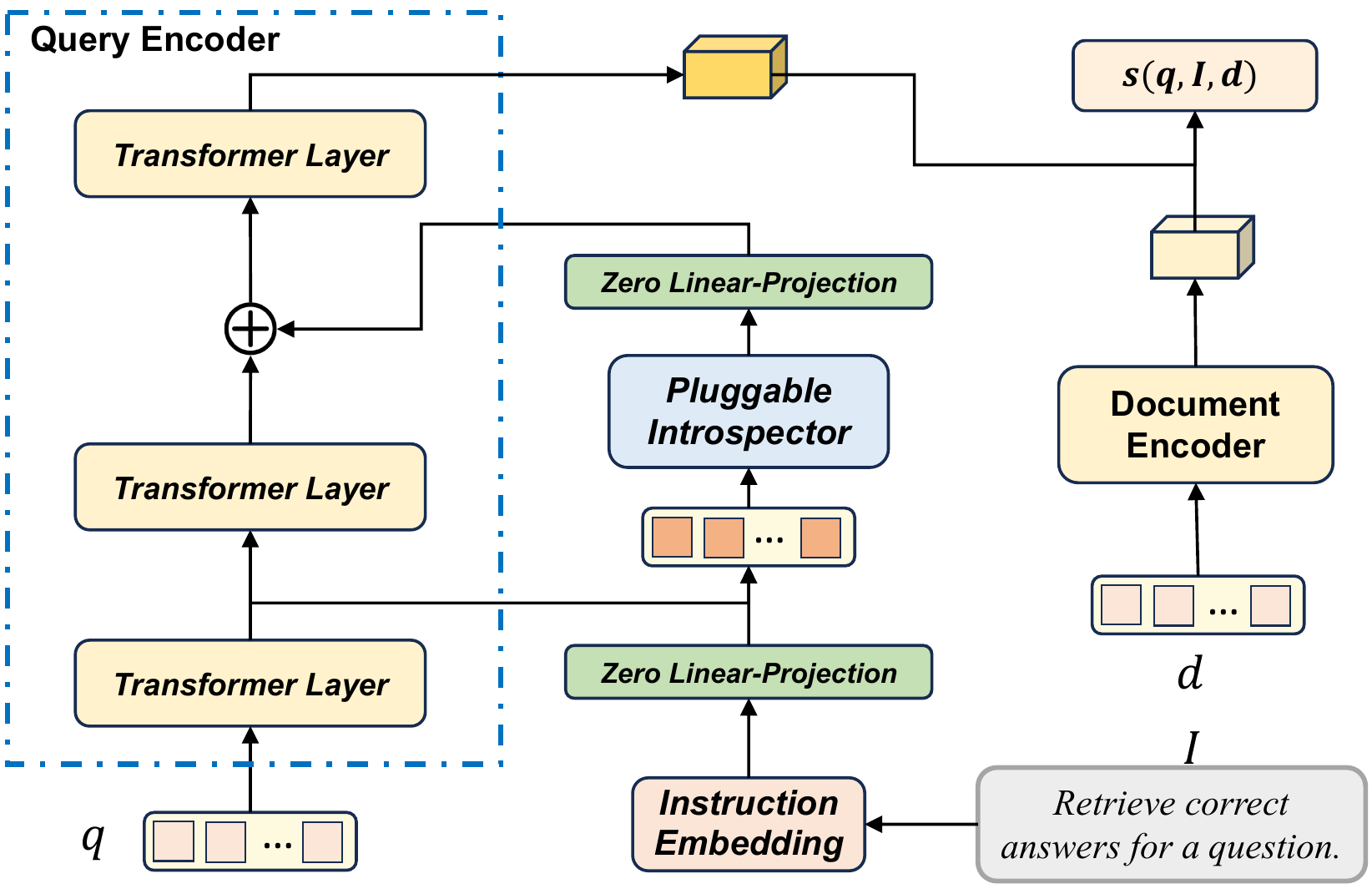}
    \caption{Parameter-isolated framework of \texttt{I3}, with a pluggable introspector enabling retrieval models to efficiently integrate the introspected intents for better query encoding.}
    \label{fig:main}
    \vspace{-1em}
\end{figure}

\begin{figure*}[t]
    \includegraphics[width=\linewidth]{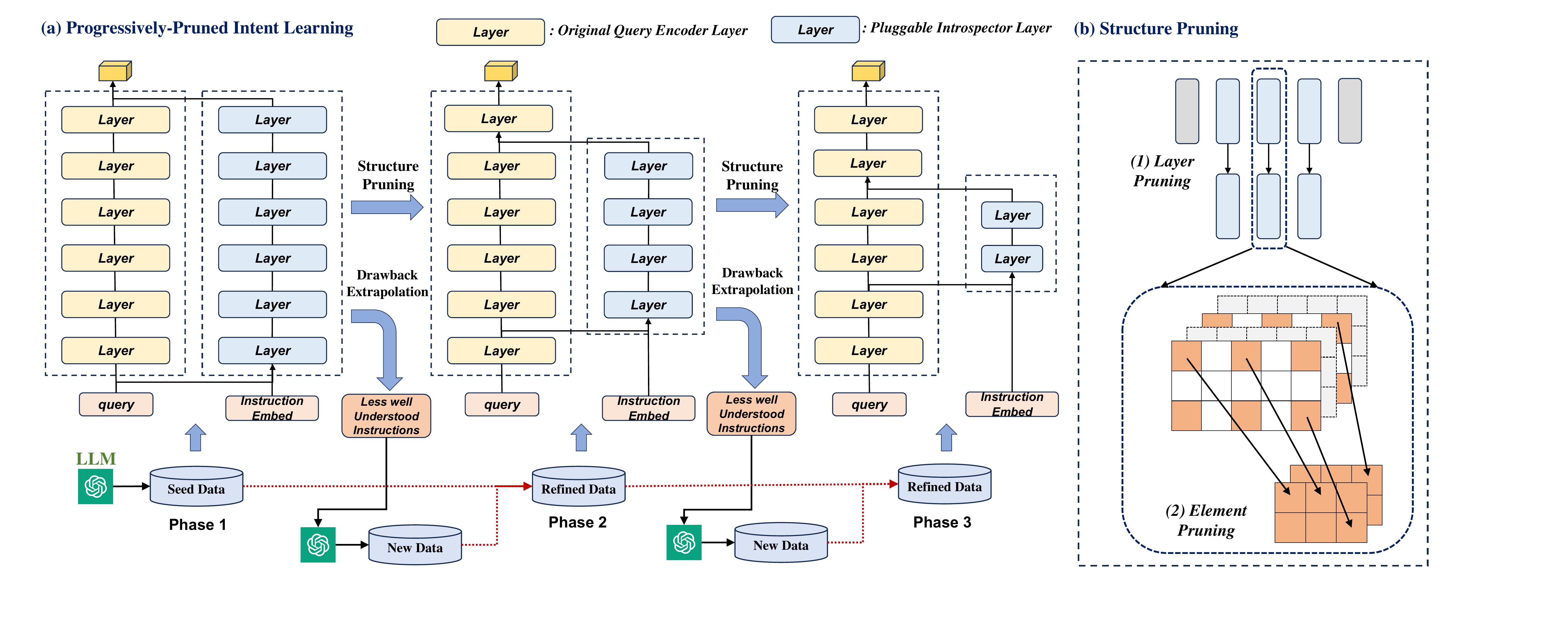}
    \vspace{-2em}
    \caption{progressively-pruned intent learning with structure pruning and drawback extrapolation-based data refinement.}\label{fig:lipper}
\vspace{-1.1em}
\end{figure*}

\subsection{Intent-Introspective Retrieval Conditioned on Instructions}
\label{sec: 3.2}
In this section, we introduce the framework of \texttt{I3}. 
As a generic approach, given an arbitrary dense retrieval model, 
\texttt{I3} fully retains its document encoder and encompasses two indispensable components within the query encoder: 
(1) a pluggable introspector, deeply comprehends the retrieval intent by jointly reasoning over the input query and instruction; 
(2) the original query encoder, which is frozen, faithfully integrates the introspected intent for better query encoding. 
These two components integrate seamlessly to construct a parameter-isolated architecture. 
This design not only preserves the inherent capability of retrieval models, but also efficiently empowers them with a new facet of intent-aware retrieval following task-specific instructions.

Specifically, we fully keep the original document encoder to eliminate the substantial time expanse of re-encoding all documents from a large corpus. 
For the query encoder, we first retain and fix its original parameters $\Theta_q$ to preserve the existing capabilities. 
Additionally, we construct a pluggable introspector (with its parameters denoted as $\Theta_p$) to explicitly introspect for specific retrieval intents.
Then the query encoding can be conceptualized as a two-step process involving \textbf{\textit{intent introspection}} and \textbf{\textit{intent-integrated encoding}}.
We first derive the query embedding from the output of the early layer within the original query encoder. 
The introspector subtly interprets the specific retrieval intent by jointly comprehending the derived query embedding and an additional instruction embedding (denoted as $c$).
The introspected intent is then seamlessly re-integrated into the late layer of the original encoder via a skip connection, enhancing subsequent encoding.

To facilitate efficient intent introspection and integration, 
we construct two instances of ``zero linear-projection" ($\mathcal{ZP}$), $\Theta_{zp1}$ and $\Theta_{zp2}$, which involve unique fully connected layers with weights and biases initialized as zeros.
Firstly, the instruction embedding $c\in \mathcal{R}^{1\times d}$ is projected to compose with the query embedding derived from the early layer's output of the original query encoder: $h_{q,c} = \mathcal{ZP}(c, \Theta_{zp1})+h^{l_{early}}_q$, 
where $h^{l_{early}}_q \in \mathcal{R}^{n\times d}$ 
and $\mathcal{ZP}(c, \Theta_{zp1})$ is added to each token of $h^{l_{early}}_q$.
And then $h_{q,c}$ serves as the input of the pluggable introspector, 
enabling it to perceive both the input query and instruction for intent comprehension.
Finally, after introspecting for the specific intents $K$, 
the intent embedding after projection is re-integrated with the hidden representations from the late layer of the query encoder, 
via skip connection: $h^{l_{late}}_q = h^{l_{late}}_q + \mathcal{ZP}(K, \Theta_{zp2})$, 
which is taken as the input to the next, \textit{i.e.}, ($l_{late}+1$)-th layer in the original encoder.
It achieves a cohesive integration between the query encoding and the introspected intents.

\paragraph{\textbf{Training Objectives. }} 
To achieve the desired outcome of intent-introspective retrieval, the two key steps, \textit{i.e.}, intent introspection, and intent-integrated encoding, should be well adapted to each other. In this regard, we design training objectives in two aspects for this mutual adaptation.

On the one hand, the intent introspection should effectively empower the subsequent query encoding. So we train the model to accurately match related queries and documents after explicitly introspecting for retrieval intents. For a given query $q_i$and corresponding instruction $\mathcal{I}_i^+$ from the training data, we select mismatched documents $\{d^-_{i,j}\}_{j=1}^m$ as negative samples, minimizing the negative log-likelihood of the relevant document $d_i^+$:
\begin{equation}
\small
\begin{aligned}
    \mathcal{L}_1 = \sum_{i=1}^{n} -log \frac{e^{s(q_i, \mathcal{I}_i^+, d_i^+)}}{e^{s(q_i, \mathcal{I}_i^+, d_i^+)} + \sum_{j=1}^m e^{s(q_i, \mathcal{I}_i^+, d_{i, j}^-)}}
\end{aligned}
\end{equation}
On the other hand, the intent-integrated encoding should also be capable of understanding the introspected intents in order to utilize the knowledge therein to the fullest extent. 
To achieve this, we explicitly optimize \texttt{I3} to recognize what constitutes correct retrieval intents in different scenarios.
For each training group of instruction, query, and relevant document, we sample some irrelevant instructions $\{\mathcal{I}^-_{i,j}\}_{j=1}^m$ as negative examples, misleading the introspector to produce incorrect retrieval intents. We optimize the negative log-likelihood of the positive instruction:
\begin{equation}
\small
\begin{aligned}
    \mathcal{L}_2 = \sum_{i=1}^{n} -log \frac{e^{s(q_i, \mathcal{I}_i^+, d_i^+)}}{e^{s(q_i, \mathcal{I}_i^+, d_i^+)} + \sum_{j=1}^m e^{s(q_i, \mathcal{I}_{i,j}^-, d_i^+)}}
\end{aligned}
\end{equation}
Finally, the training loss is represented as $\mathcal{L} = \mathcal{L}_1 + \alpha \mathcal{L}_2$, with $\alpha$ as the hyper-parameter. During training, we fix the parameters of the original dual-encoders (\textit{i.e.}, $\Theta_q$ and $\Theta_d$), only optimizing parameters of the pluggable introspector and the two instances of zero linear-projection (\textit{i.e.}, $\Theta_p$, $\Theta_{zp1}$ and $\Theta_{zp2}$). This realizes efficient training while simultaneously effectively preserving the original capabilities of retrieval models.

\paragraph{\textbf{Analysis of Harmless Intent Integration. }} 
Because the weight and bias of the fully connected layers in ``zero linear-projection'' are both initialized as zeros, it ensures harmless intent integration, providing better optimization than training from scratch for the introduced parameters. 
Specifically, in the initial training step, the instruction embedding $c$ does not introduce any additional noise to the query embedding.
Moreover, the integration of intent embedding also does not alter the hidden representations within the original query encoder, effectively maintaining the original output as if the pluggable introspector did not exist. 
It indicates that before any parameter optimization, integrating the pluggable introspector into the retrieval model is completely harmless to the original encoding process.
With the parameters of the original dual-encoders fixed, we \textit{ensure the preservation of previously learned knowledge, while also facilitating efficient intent introspection and integration}.  
Consequently, \texttt{I3} effectively enhances retrieval models, controlling them to perform intent-aware retrieval following instructions.

\subsection{Progressively-Pruned Intent Learning}
\label{sec: 3.3}
To efficiently perform intent-introspective retrieval following instructions, the pluggable introspector should be trained to adeptly perceive various retrieval intents within instructions, while also maintaining a lightweight architecture to guarantee minimal impact on the time efficiency of query encoding.
To achieve this goal, we propose \textbf{progressively-pruned intent learning}, as shown in Figure~\ref{fig:lipper}. 
It harnesses the advancement of LLMs to automatically generate extensive data with instructions as the seed training dataset, and then divides the total training into several phases. 
In each phase, we leverage the LLM-generated data to train \texttt{I3} with the objectives mentioned in \S \ref{sec: 3.2}, and also incorporate two key designs: 
(1) \textbf{\textit{progressive structure pruning}} that streamlines the pluggable introspector into a more lightweight module; 
(2) \textbf{\textit{drawback extrapolation-based data refinement}} that extrapolates the drawback of the current model to synthesize additional data for enhancing the initial seed training dataset.

\subsubsection{\textbf{LLM-guided Instruction Data Synthesizing}}
\label{sec: 3.3.1}
To effectively optimize \texttt{I3} for intent introspection conditioned on instructions, first it is crucial to construct a diverse set of training data comprising various retrieval instructions. We develop a generation pipeline that utilizes an LLM to automatically synthesize a large amount of query-document pairs together with instructions as the seed training dataset.

\textbf{(1) Instruction Generation.} 
First, we prompt the LLM to generate a diverse range of retrieval instructions. 
To ensure comprehensive expression of different retrieval tasks, when generating instructions, we require the LLM to specify the topic (\textit{e.g.}, scientific, legal) and organizational formats (\textit{e.g.}, sentence, paragraph, dialogue) of the retrieved text, while also incorporating a clear definition of relevance (\textit{i.e.}, search intent) for the retrieval task. 
Moreover, To further elevate the quality of instruction generation, we also incorporate some instruction examples (selected from previously generated instructions) into the prompt template as in-context samples.
\textbf{(2) Query-Document Pair Generation. } 
For each generated instruction, we subsequently ask the LLM to write some appropriate documents and their associated queries. The generated documents and queries should align with the designated topic, organizational formats and the relevance definition outlined in the instructions.
\textbf{(3) Query Self-check. } However, the query and document simultaneously generated in the preceding step may not always correctly capture the retrieval intents expressed in the instructions. To tackle this issue, we ask the LLM to verify if the query-document relationship matches the relevance criteria set by the instruction. And the LLM is further required to rewrite the query that fails to meet the relevance criteria, ensuring a cohesive association among the instruction, query, and document.

Through the above three steps, we have generated extensive seed data, encompassing a diverse range of retrieval instructions.
For each instruction, we select a small subset of corresponding query-document pairs to form the validation set, while the remainder is utilized as the seed training dataset for fine-tuning \texttt{I3}.

\subsubsection{\textbf{Progressive Structure Pruning}}
\label{sec: 3.3.2}

To design the specific structure of the pluggable introspector for efficient intent introspection, a straightforward approach entails copying an extra query encoder as the introspector and training it to understand diverse instructions\citep{controlnet}.
However, the introspector only needs to perceive specific intents over the input query and instruction, without the necessity to match the model size of the query encoder. 
Moreover, it is essential for the introspector to maintain a lightweight design, ensuring minimal impact on the time efficiency of query encoding.

To address these considerations, we duplicate an additional query encoder as the initial pluggable introspector, and propose \textbf{\textit{progressive structure pruning}} that prunes the introspector into a sheared structure after each training phase. 
Only a subset of the weights from the larger introspector is selected to initialize the smaller version. 
This facilitates the transfer of knowledge learned by the larger introspector to the smaller counterpart, ensuring that the model becomes increasingly lightweight with almost no performance degradation.

Specifically, prior to the first training phase, the pluggable introspector possesses the same architecture and parameters as the original query encoder. 
In this setup, the introspector and the original query encoder are connected only at their input and output spaces: the input embedding of the query encoder is derived as the input for the introspector, and the introspected intent embedding is directly integrated with the output of the query encoder, producing the final query representation.
In subsequent phases, before training we first perform structure pruning on the introspector from the previous phase, streamlining it into a sheared module. 
And the weights of the sheared introspector are initialized from the larger counterpart, resembling the process of knowledge distillation. 
We refer to the larger introspector before pruning as the \textit{teacher introspector}, and the target sheared one as the \textit{student introspector}.
And structure pruning involves two aspects: layer pruning and element pruning, as shown in Figure~\ref{fig:lipper}.b. 

\textbf{\textit{(1) Layer pruning: }} 
The student introspector comprises a reduced number of transformer layers compared to its teacher counterpart. And each layer in the student introspector is initialized using a corresponding layer from the teacher. 
Specifically, we omit several initial and final layers while retaining the intermediate ones from the teacher introspector.
Consequently, the early derivation of query embedding which serves as the input for the introspector, and the late integration of specific retrieval intents, both take place within the more intermediate layers of the original query encoder.
This not only enables the query encoder to provide enhanced query embeddings for the introspector, but also allows more subsequent layers in the query encoder to effectively integrate the introspected intents for improved encoding.

\textbf{\textit{(2) Element pruning: }} 
After layer pruning, we need to initialize each component within the student introspector's layers, using the corresponding larger counterpart from the teacher introspector.
To streamline the model architecture, targets for element pruning may include the number of attention heads, and the hidden or intermediate dimension within the transformer layer, among others.
We employ a uniform selection strategy, wherein evenly-spaced elements are selected from the teacher's tensor to initialize the student's corresponding component.
For example, when leveraging weight tensor $W_t \in \mathcal{R}^{t_1\times t_2\times \ldots \times t_n}$ from the teacher introspector to initialize the student's weight tensor $W_s \in \mathcal{R}^{s_1\times s_2\times \ldots \times s_n}$ which is of the same component type ($s_i \leq e_i$), we evenly-spaced select $s_i$ slices out of $t_i$  for each dimension $i$ of $W_t$ to facilitate initialization of $W_s$.
And previous research~\citep{xu2023initializing} has demonstrated that such a uniform selection strategy is likely to yield benefits of knowledge transfer from the teacher model to the student.

Besides pruning the introspector into a sheared structure, at the beginning of each training phase, we also reinitialize the weights and biases within the two ``zero linear-projection" instances to zeros. 
Throughout each phase of training, we consolidate and strengthen the model capability based on that inherited from the larger introspector in the preceding phase.
Ultimately, we streamlined the introspector into a more lightweight module, which improves the time efficiency of query encoding while maintaining the performance with minimal degradation.

\subsubsection{\textbf{Drawback Extrapolation-Based Data Refinement}}
\label{sec: 3.3.3}
We then describe the drawback extrapolation-based data refinement. It extrapolates the drawbacks of the trained model from each training phase by detecting instructions that the model struggles to understand, and then specifically generates new training data to enhance the original seed dataset.

In particular, after each training phase, we evaluate the current model's performance on the synthesized validation set for each instruction. 
Instructions that are not fully comprehended by the model inevitably result in sub-optimal performance.
We then utilize these challenging instructions as in-context samples for the LLM to generate new instructions, which often bear a resemblance to the original in-context instructions and may similarly pose comprehension challenges to the current model.
Moreover, we follow the data-generation pipeline to synthesize corresponding query-document pairs that align with these new instructions, and incorporate them into the original training dataset for refinement. 
In subsequent training phases, these data can specifically target the identified weaknesses of the current model, thereby enhancing its ability to understand a diverse range of retrieval instructions.

\section{Experimental Setup}

To illustrate the effectiveness of \texttt{I3}, we evaluate our approach in two settings: (1) zero-shot evaluation on retrieval scenarios and (2) zero-shot evaluation on reranking scenarios after cooperating \texttt{I3} with reranking models.

\subsection{Benchmark} Our experiments are conducted on the BEIR~\citep{beir} benchmark for zero-shot evaluation on retrieval and reranking scenarios.
Following prior works\citep{beir, contriever}, we leverage Normalized Discounted Cumulative Gain 10 (nDCG@10) as the evaluation metric and use the 15 publicly available datasets in BEIR for evaluating retrieval models, including 14 out-of-domain datasets (\textit{i.e.}, \textbf{TREC-COVID}~\citep{trec-covid}, \textbf{NFCorpus}~\citep{nfcorpus}, \textbf{NQ}~\citep{nq}, \textbf{HotpotQA}~\citep{hotpotqa}, \textbf{FiQA-2018}~\citep{fiqa}, \textbf{ArguAna}~\citep{arguana}, \textbf{Touché-2020}~\citep{touche}, \textbf{Quora}~\citep{beir}, \textbf{DBPedia-entity}~\citep{dbp}, \textbf{SCIDOCS}~\citealp{scidocs}, \textbf{Fever}~\citep{fever}, \textbf{Climate-Fever}~\citep{climate}, \textbf{SciFact}~\citep{scifact}, \textbf{CQADupStack}~\citep{cqa}) 
and one in-domain dataset (\textit{i.e.}, \textbf{MS MARCO}~\citep{msmarco}). 
Besides, the evaluation instructions for all datasets, which describe the retrieval intents, are in alignment with those employed in \cite{tart}.
Moreover, When evaluating reranking models, we conduct experiments across 11 of the above datasets following \citep{promptagator}.

\subsection{Baselines} 
\paragraph{\textbf{Retrieval Baselines. }} We compare \texttt{I3} with various competitive retrieval methods, which can be categorized into four groups. 
\textit{The first group} employs a sparse retrieval method known as \textbf{BM25}. 

\textit{The second group} trains retrievers on a few supervised datasets (\textit{e.g.}, MS MARCO~\citep{msmarco}) and directly transfer them to new tasks, 
including \textbf{Contriever}~\citep{contriever}, \textbf{GTR}~\citep{gtr}, \textbf{ColBERT-v2}~\citep{colbertv2}, \textbf{CPT}~\citep{cpt}, \textbf{LaPraDoR}~\citep{laprador}, \textbf{COCO-DR}~\citep{cocodr}, and \textbf{SGPT}~\citep{sgpt}. 
These baseline models have varying parameter sizes. Notably, GTR comprises 4.8B parameters, SGPT consists of 5.8B parameters, and CPT boasts an impressive 175B parameters, almost directly utilizing LLMs for retrieval tasks.

\textit{The third group} of models train a task-specific retriever for each downstream task with pseudo generated training data, including \textbf{GenQ}~\citep{beir}, \textbf{GPL}~\citep{gpl}, \textbf{Promptgator}~\citep{promptagator}. 
GenQ and GPL utilize a specifically trained T5 for data generation. And Promptgator is a few-shot LLM-enhanced method that prompts LLMs to synthesize training data.

And \textit{the final group} includes \textbf{TART-dual}~\citep{tart} and \textbf{InstructOR}~\citep{su-etal-2023-one}. TART-dual collects a large number of supervised datasets and augments them with human-annotated instructions. The collected datasets, known as $Berri$, are then employed to train a single retriever to solve different tasks with instructions. And InstructOR adopts a similar approach, focusing on instruction-based retrieval.

\paragraph{\textbf{Reranking Baselines.}} In contexts where speed is not critical, the reranking model with a cross-encoder is often used to compute the query-document relevance by jointly encoding them with cross-attention. 
Under the reranking settings, a retrieval model is first utilized to retrieve the Top-K documents, followed by the application of a reranking model to reorder these retrieved documents.  
In our experimental setup, we first leverage \texttt{I3} to retrieve the Top-K documents (where K=100) and then employ a competitive reranking model, monoT5 (3B)~\citep{monot5}, to reorder these retrieved documents. 

We compare with the following state-of-the-art Retriever+Reranker combinations:
\textbf{UPR (3B)}~\cite{upr} that initially retrieves 1000 documents with Contriever, 
\textbf{Contriever+CE} that initially retrieves 100 documents with Contriever, 
\textbf{BM25+monoT5} (3B)~\citep{monot5, monot5_bm25} that initially retrieves 100 or 1000 documents with BM25,
\textbf{COCO-DR+monoT5} (3B)~\citep{monot5, monot5_bm25} that initially retrieves 100 documents with COCO-DR,
\textbf{Promptgator++} (zero-shot version \& few-shot version)~\citep{promptagator} that initially retrieves 200 documents with Promptgator, 
\textbf{TART-full} (T0-3B version \& Flant5-XL version)~\citep{tart} that initially retrieves 100 documents with Contriever.

\begin{table*}[t]
    \centering
    \caption{\label{tab:main_result}nDCG@10 on the BEIR Benchmark. We compare \texttt{I3} with other retriever models. Avg CPT Sub is the average performance on 11 BEIR tasks used in \citep{cpt}. Avg TART Sub is the average performance on 9 BEIR tasks used in \citep{tart}.}
    \vspace{-1em}
    \begin{adjustbox}{width=\textwidth}
    \begin{tabular}{l|c|ccccccc|ccc|cc|c}
    \toprule
   \textbf{Datasets}  & BM25 & Contriever & GTR & ColBERTv2 & CPT & LaPraDoR & COCO-DR & SGPT & GenQ & GPL & Promptgator & TART-dual & InstructOR & \textbf{\texttt{I3}}\\
     \midrule
MS MARCO &  {22.8}  &  {40.7} &  {\bf{44.2}}     &  {---} &  {---} &  {36.6}  &  {42.4}   &  {39.9} &  {40.8} &  {---} &  {---}        &  {---} & {41.6} & {41.8} \\
\midrule
TREC-COVID    &  {65.6}  &  {59.6} &  {50.1}     &  {73.8} &  {64.9} &  {77.9}  &  {80.4}   &  \bf{87.3} &  {61.9} &  {70.0} &  {75.6}        &  {62.6}& {71.4} & {81.6} \\
NFCorpus    &  {32.5}  &  {32.8} &  {34.2}     &  {33.8} &  {\bf{40.7}} &  {34.7}  &  {35.4}   &  {36.2} &  {31.9} &  {34.5} &  {33.4}        &  {33.7}& {36.0} & {37.1} \\
NQ       &  {32.9}  &  {49.8} &  {56.8}     &  {56.2} &  {---} &  {47.9}  &  {54.7}   &  {52.4} &  {35.8} &  {48.3} &  {---}        &  {---} & {57.3}& {\bf{57.4}} \\
HotpotQA       &  {60.3}  &  {63.8} &  {59.9}     &  {66.7} &  {\bf{68.8}} &  {66.6}  &  {64.1}   &  {59.3} &  {53.4} &  {58.2} &  {61.4}        &  {---} & {55.9}& {63.3} \\
FiQA-2018    &  {23.6}  &  {32.9} &  {46.7}     &  {35.6} &  {\bf{51.2}} &  {34.3}  &  {32.9}   &  {37.2} &  {30.8} &  {34.4} &  {46.2}        &  {33.7}& {47.0} & {35.7} \\
ArguAna        &  {41.4}  &  {44.6} &  {54.0}     &  {46.3} &  {43.5} &  {50.8}  &  {51.5}   &  {51.4} &  {49.3} &  {55.7} &  {59.4}   &  {48.9} & {55.7}& {\bf{59.8}} \\
Touché-2020     &  {\bf{36.7}}  &  {23.0} &  {25.6}     &  {26.3} &  {29.1} &  {33.3}  &  {26.3}   &  {25.4} &  {18.2} &  {25.5} &  {34.5}        &  {20.1}& {23.4} & {23.7} \\
Quora     &  {78.9}  &  {86.5} &  {89.2}  &  {85.2} &  {63.8} &  {87.5}  &  {87.2}   &  {84.6} &  {83.0} &  {83.6} &  {---}        &  {---} & {88.9}& {\bf{89.3}} \\
DBPedia-entity    &  {31.3}  &  {41.3} &  {40.8}   &  {\bf{44.6}} &  {43.2} &  {39.1}  &  {40.7}   &  {39.9} &  {32.8} &  {38.4} &  {38.0}        &  {41.5}& {40.2} & {41.8} \\
SCIDOCS     &  {15.8}  &  {16.5} &  {16.1}     &  {15.4} &  {---} &  {18.4}  &  {17.8}   &  {19.7} &  {14.3} &  {16.9} &  {18.4}        &  {14.2}& {17.4} & {\bf{19.9}} \\
Fever      &  {75.3}  &  {75.8} &  {74.0}  &  {78.5} &  {77.5} &  {76.3}  &  {79.3}   &  {78.3} &  {66.9} &  {75.9} &  {77.0}        &  {---}& {70.0} & {\bf{80.8}} \\
Climate-Fever    &  {21.3}  &  {23.7} &  {26.7}     &  {17.6} &  {22.3} &  {26.1}  &  {24.7}   &  {30.5} &  {17.5} &  {23.5} &  {16.8}        &  {13.8}& {26.5} & {\bf{31.1}} \\
SciFact   &  {66.5}  &  {67.7} &  {66.2}   &  {69.3} &  {75.4} &  {68.7}  &  {72.2}   &  {74.7} &  {64.4} &  {67.4} &  {65.0}        &  {69.0} & {64.6}& {\bf{79.9}} \\
CQADupStack   &  {29.9}  &  {34.5} &  {39.9}     &  {---} &  {---} &  {29.0}  &  {39.3}   &  {38.1} &  {34.7} &  {35.7} &  {---}        &  {---}& {\bf{43.0}} & {40.0} \\ \midrule
\bf Avg CPT(TART) Sub            &  {48.5}  &  {50.2} &  {51.6}     &  {52.5} &  {52.8} &  {54.1}  &  {54.1}   &  {55.0} &  {46.4} &  {51.6} &  {---(43.0)}        &  {---(37.4)} &{52.7}& {\bf{56.7(45.6)}} \\
\bf Avg             &  {43.7}  &  {46.6} &  {48.6}     &  {---} &  {---} &  {49.3}  &  {50.5}   &  {51.1} &  {42.5} &  {47.7} &  {---}        &  {---} & {49.8}& {\bf{52.9}} \\
    \bottomrule
    \end{tabular}
    \end{adjustbox}
    \vspace{-1em}
\end{table*}
\begin{table*}[t]
    \centering
    \caption{\label{tab:reranking}We cooperate \texttt{I3} with monoT5~\cite{monot5} and compare with other reranking baselines on 11 datasets in BEIR.}
    \vspace{-1em}
    \begin{adjustbox}{width=\textwidth}
    \begin{tabular}{l|c|ccccccccccc|cc}
    \toprule
    \textbf{Methods} & K & arg & touché & covid & nfc & hotpot & dbp & climate & fever & scifact & scidocs & fiqa & \bf{Avg TART Sub}  & \bf{Avg} \\  \midrule
    UPR (3B) & 1000 & {50.3} & {21.3} & {60.4} & {33.3} & {72.2} & {33.8} &{9.5} & {57.3} & {69.6} & {17.3} & {45.0} &{37.8} & {42.7} \\ \midrule
    Contriever+CE & 100 & {41.3} & {29.8} & {70.1} & {34.4} & {---} & {47.1} &{25.8} & {---} & {69.2} & {17.1} & {36.7} &{41.3} & {---} \\ 
    BM25+monoT5 (3B) & 100 &{32.3} & {21.1} & {82.0} & {39.4} & {73.1} & {44.1} & {27.0} & {83.9} & {76.2} & {19.4} & {46.2} & {43.0} & {49.5} \\
    BM25+monoT5 (3B) &1000&{38.0} &{30.0} & {79.5} &{38.4} &{\textbf{75.9}} &{47.8} &{28.0} &{85.0} &{77.7} & {19.7} &{51.4} &{45.6} &{51.9} \\
    COCO-DR+monoT5 (3B) & 100 &{40.6} & {28.4} & {85.9} & {39.8} & {71.4} & {47.7} & {29.1} & {84.6} & {76.4} &{20.0} & {48.5} & {46.3} & {52.0} \\ 
    Promptgator++ (zero-shot) & 200 & {52.1} & {27.8} & {76.0} & {36.0} & {71.2} & {41.3} & {22.6} & {83.8} & {73.2} & {19.1} & {45.9} &{43.8} &{49.9}\\
    Promptgator++ (few-shot) & 200 & {\textbf{63.0}} & {38.1} & {76.2} & {37.0} & {73.6} & {43.4} & {20.3} & {86.6} & {73.1} & {20.1} & {49.4} &{46.7} &{52.8}\\ \midrule
    TART-full (T0-3B) & 100 & {49.8} &{\textbf{31.2}} & {71.7} & {34.0} & {---} &{45.1}&{30.0}&{---} & {75.8} & {17.5} & {42.2} & {44.1} & {---}\\
    TART-full (FlanT5-XL) & 100 & {51.5} &{24.9} & {72.8} & {33.4} & {---} &{46.8} & {35.4} &{---}& {77.7} & {18.7} & {41.8} & {44.8} & {---}\\ \midrule
    \textbf{\texttt{I3}+monoT5(3B)} & 100 &{41.4} & {29.0} & {\textbf{86.3}} & {\textbf{40.0}} & {72.8} & {\textbf{48.5}} & {\textbf{35.6}} & {\textbf{86.7}} & {\textbf{77.9}} &{\textbf{20.4}} & {\textbf{51.4}} & {\bf{47.8}} & {\bf{53.6}} \\
    
    \bottomrule
    \end{tabular}
    \end{adjustbox}
    \vspace{-0.9em}
\end{table*}
\subsection{Implementation Details}
To automatically generate retrieval data with diverse instructions, we leverage ChatGPT~\citep{chatgpt} (OpenAI \texttt{gpt-3.5-turbo}) as the LLM. 
We set the temperature to 1 and generate approximately 100 instructions with different retrieval intents.
Concurrently, we synthesize about 140K query-document pairs in total, including 100K pairs in the seed training dataset and 20K pairs created during data refinement.
Furthermore, we rigorously eliminate the generated instructions that exhibit similar retrieval intents as downstream tasks, thereby \textit{\textbf{preventing any overlap or similarities between the generated training data and the test data.}}

With LLM-generated training data, we leverage COCO-DR$_{\text{Large}}$~\citep{cocodr}, a dual-encoder consisting of 335M parameters, as the backbone model to implement \texttt{I3}. Specifically, COCO-DR$_{\text{Large}}$ has the same architecture as BERT$_{\text{Large}}$~\citep{bert} 
and takes the [CLS] pooling of the top encoder layer as the query/document embeddings. 
We retain the original dual-encoders and duplicate an extra query encoder as the pluggable introspector. 
And the instruction embedding $c$ is pre-encoded by the original query encoder.
Moreover, we encompass three phases within progressively-pruned intent learning
and perform structure pruning after each of the first two phases. In the final version of the introspector, the number of layers is reduced from the original 24 to 12, the hidden dimension decreases from 1024 to 768, the intermediate dimension decreases from 4096 to 3072, and the number of attention heads decreases from 16 to 12.

During training, we set the batch size as 64, with the maximum sequence length as 256, the learning rate as 5e-6, and $\alpha$ as 0.5. When conducting contrastive learning, we randomly sample 4 mismatched instructions as negative examples in $\mathcal{L}_2$. Moreover, in each phase we leverage the model from the previous phase (COCO-DR in the initial phase) to retrieve documents with high similarity scores yet unrelated to the query from generated documents as hard negatives, together with in-batch negatives for $\mathcal{L}_1$. Furthermore, when evaluating on BEIR, we follow the same hyperparameters in COCO-DR~\citep{cocodr} to ensure a fair comparison.

\section{Experimental Results}

\subsection{Main Results on Retrieval Scenarios}
Table~\ref{tab:main_result} lists the zero-shot evaluation results of retrieval models on BEIR. The results for baseline methods are derived from their respective papers. As some baselines employ datasets from BEIR for training, their downstream test excludes these datasets to ensure a fair comparison, leading to results missing for some baselines on specific datasets. When compared with these baselines, we compute the average performance for \texttt{I3} exclusively on the datasets they test. We have the following observations from the experimental results.

\textbf{First of all}, \texttt{I3} outperforms all previous retrieval models on the average nDCG@10 metric of BEIR tasks. It shows significant improvements over its backbone model, \textit{i.e.}, COCO-DR, improving the average nDCG@10 by 2.3 points and achieving superior performance on 12 of 14 out-of-domain tasks. The significantly higher zero-shot performance of our model demonstrates the stronger generalizability of \texttt{I3} across various retrieval tasks.

\textbf{Second}, while some approaches (\textit{e.g.}, Promptgator) leverage LLMs to generate pseudo training data and train task-specific models for each downstream task, our model achieves superior transferability across diverse retrieval intents of tasks. For example, \texttt{I3} yields 2.6 point nDCG improvement over the few-shot method Promptgator. This indicates that our \texttt{I3} can efficiently and effectively transfer to different retrieval tasks using only instructions without any task-specific training.

\textbf{Third}, compared with models that have significantly more parameters,\textit{e.g.}, SGPT (\textbf{5.8B} parameters), CPT (\textbf{175B}), our method still achieves stronger average performance with much fewer parameters (\textbf{445M}). 
It highlights smaller models can still achieve remarkable effectiveness when designed with an ingenious architecture to well understand the retrieval intents.

\textbf{Finally}, \texttt{I3} leverages a large margin improvement compared with TART-dual (45.6 v.s. 37.4 on average) and InstructOR (52.9 v.s. 49.8 on average), which leverage human-annotated training data for instruction-based retrieval. 
We argue that it can be attributed to two key factors. 
First, the parameter-isolated architecture of \texttt{I3} better preserves the original capability of the retrieval model, while also flexibly unleashing its ability to conduct intent-introspective retrieval following instruction.
Second, progressively-pruned intent learning explicitly optimizes the transferability of \texttt{I3} based on extensive automatically LLM-generated retrieval data with diverse search intents.

\newcommand{\tabincell}[2]{\begin{tabular}{@{}#1@{}}#2\end{tabular}} 
\begin{table}[t]
    \centering
    \caption{\label{tab:ablation}Results of ablation study to illustrate the effect of individual components.}
    \vspace{-1em}
    \begin{adjustbox}{width=0.48\textwidth}
    \begin{tabular}{l|c|cccc|cc}
    \toprule
     & \tabincell{c} {\textbf{0} backbone\\COCO-DR} & \tabincell{c} {\textbf{1} w/o \textit{intro-} \\ \textit{spection}} &  \tabincell{c} {\textbf{2} w/o \\ \textit{instr}} & \tabincell{c} {\textbf{3} w/o \\\textit{progressive}}
     & \tabincell{c} {\textbf{4} w/o \\\textit{refine}}& \tabincell{c} {\textbf{5} w/o \\\textit{pruning}}  & \textbf{\texttt{I3}}\\
     \midrule
MS MARCO &  {42.4} &  {37.9}  &  {40.3}  &  {40.9}  &  {41.7}  &  {41.8}   &  {41.8} \\
     \midrule
TREC-COVID &  {80.4} &  {77.9}  &  {76.1}  &  {80.3}  &  {80.8}  &  {\bf{82.1}} &  {81.6}\\
NFCorpus &  {35.4} &  {34.1}  &  {36.2}  & {36.3}  &  {36.4}  &  {36.5}  &  {\bf{37.1}} \\
NQ  &  {54.7} &  {51.0}  &  {55.3} &  {56.0}  &  {56.3}  &  {57.2} &  {\bf{57.4}} \\
HotpotQA  &  {64.1} &  {58.6}  &  {61.0} &  {62.9}  &  {63.2} &  {\bf{64.5}}  &  {63.3} \\
FiQA-2018  &  {32.9} &  {31.8}  &  {34.1}  & {34.8}  &  {35.2}  &  {35.3} &  {\bf{35.7}}  \\
ArguAna  &  {51.5} &  {56.4}  &  {55.9}  &  {57.4}  &  {56.4}  &  {58.8}&  {\bf{59.8}} \\
Touché-2020&  {\bf{26.3}} &  {25.5}  &  {22.8}  &  {22.9}  &  {24.3}  &  {24.5}  &  {23.7}\\
Quora &  {87.2} &   {86.7}  &  {87.6}  & {87.7}  &  {88.2}  &  {88.6}  &  {\bf{89.3}}\\
DBPedia-entity &  {40.7} &   {32.3}  &  {41.2} &  {41.4}  &  {41.0}  &  {\bf{42.1}} &  {41.8}\\
SCIDOCS &  {17.8}&   {16.7}  &  {18.2} & {18.8}  &  {19.2} &  {19.6} &  {\bf{19.9}} \\
Fever &  {79.3} &   {77.7}  &  {79.7}  & {80.1}  &  {80.4}  &  {\bf{81.3}}  &  {80.8}\\
Climate-Fever &  {30.4} &   {25.3}  &  {30.9} &  {30.6}  &  {30.9} &  {\bf{32.0}}  &  {31.1} \\
SciFact  &  {72.2} &   {72.0}  &  {77.4}  &  {78.5}  &  {78.8}  &  {79.0} &  {\bf{79.9}}\\
CQADupStack &  {39.3} &   {38.3}  &  {39.5}  &  {39.1}  &  {39.3}  &  {\bf{40.1}} &  {40.0} \\    \midrule
\bf Avg  &  {50.5}&   {48.9}  &  {51.1}  &  {51.9}  &  {52.2} &  {\bf{53.0}} &  {52.9} \\
    \bottomrule
    \end{tabular}
    \end{adjustbox}
    \vspace{-1.1em}
\end{table}

\begin{table}[t]
    \centering
    \caption{\label{tab:effiency}The average retrieval time efficiency across each dataset within BEIR.}
    \vspace{-1em}
    \small
    \begin{adjustbox}{width=0.45\textwidth}
    \begin{tabular}{lc|cc}
    \toprule
    \textbf{Method} & \textbf{COCO-DR} & \textbf{\texttt{I3} w/o pruning} & \textbf{\texttt{I3}} \\ \midrule
    \#Query per second & 1600 & 700 & \bf{1150}  \\
    \#Document per second  & 500 & 500 & 500  \\
    Query encoding time& 2.3s & 5.2s & \bf{3.1}s \\ 
    Document encoding time & 4514s & 4514s & 4514s \\ 
    Retrieval Latency &  7.5s & 7.5s & 7.5s\\ \midrule
    Total Time & 4523.8s & 4526.7s & \bf{4524.6}s\\
     \bottomrule
    \end{tabular}
    \end{adjustbox}
    \vspace{-1em}
\end{table}

\subsection{Main Results on Reranking Scenarios}
To further illustrate that our method can be applied to reranking scenarios and yield performance improvements, we cooperate \texttt{I3} with a competitive reranking model, monoT5 (3B)~\citep{monot5} and conduct experiments on BEIR. 
As shown in \textbf{Table~\ref{tab:reranking}}, the combination of \texttt{I3}+monoT5 has achieved the new state-of-the-art reranking performance in BEIR. 
Compared to BM25+monoT5 and COCO-DR+monoT5, our approach involves a mere substitution of the retrieval model, yet results in a marked enhancement in performance.
And the combination of \texttt{I3}+monoT5 is also notably superior to zero-shot reranking models like TART-full~\citep{tart} and Promptgator++~\citep{promptagator}. 
With a zero-shot setup, it even outperforms few-shot Promptgator++ which achieves previous SOTA performance, exhibiting average gains of 0.8 nDCG@10 points. 

Furthermore, for certain baseline methods like BM25+monoT5 and Promptgator++, we adhere to the experimental settings in their original papers, and a value larger than 100 is selected for $K$ as the number of documents initially retrieved.
Typically, A larger K implies a greater reliance on the capabilities of more powerful reranking models that utilize cross-encoders. 
It tends to improve the reranking performance but also increases the time cost for reranking (the time cost is proportional to the value of K), as reranking is generally more time-consuming compared to retrieval.  
In contrast to baselines with larger K values, our method still achieves superior performance under more stringent conditions and additionally reduces the time cost of reranking.

\subsection{In-Depth Analysis}

\paragraph{\textbf{Effect of Individual Components. }} 
We conduct the ablation study to illustrate the effect of each component in \textbf{Table~\ref{tab:ablation}}. Specifically, we train the following ablation models:
(1) \textbf{\textit{w/o introspection}}: we remove the pluggable introspector and  directly tune the COCO-DR$_{\text{Large}}$ backbone with our generated data, following TART-dual. 
(2) \textbf{\textit{w/o instr}}: we do not provide any instructions to \texttt{I3} during both training and evaluation.
(3) \textbf{\textit{w/o progressive}}: Before training, we prune the original query encoder directly to the desired sheared structure to serve as the pluggable introspector, no longer conducting progressive structure pruning during subsequent training phases.
(5) \textbf{\textit{w/o refine}}: We do not perform drawback extrapolation-based data refinement, with the synthesized seed dataset as the only training data.
(4) \textbf{\textit{w/o pruning}}: We do not perform structure pruning and the pluggable introspector always maintains the same structure as the original query encoder.

\begin{figure}[t]
    \includegraphics[width=\linewidth]{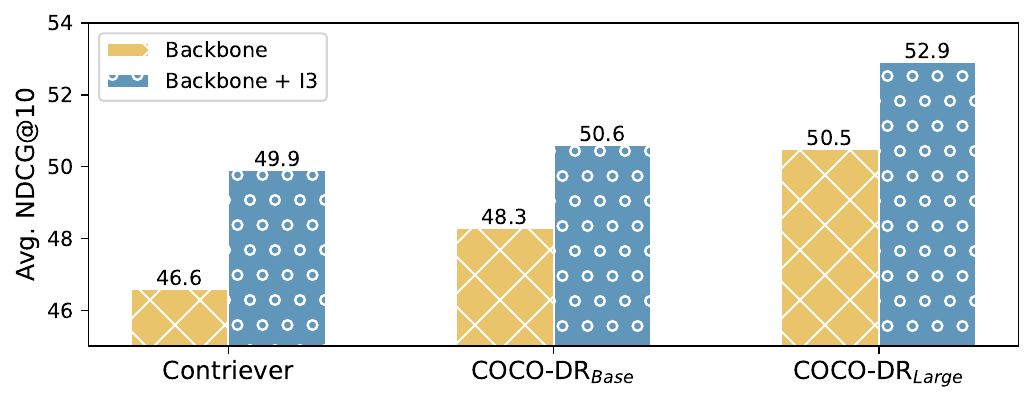}
    \vspace{-2.4em}
    \caption{Average nDCG@10 of \texttt{I3} incorporating different backbones on the BEIR benchmark.}\label{fig:abla1}
\vspace{-0.8em}
\end{figure}

\begin{figure}[t]
    \includegraphics[width=\linewidth]{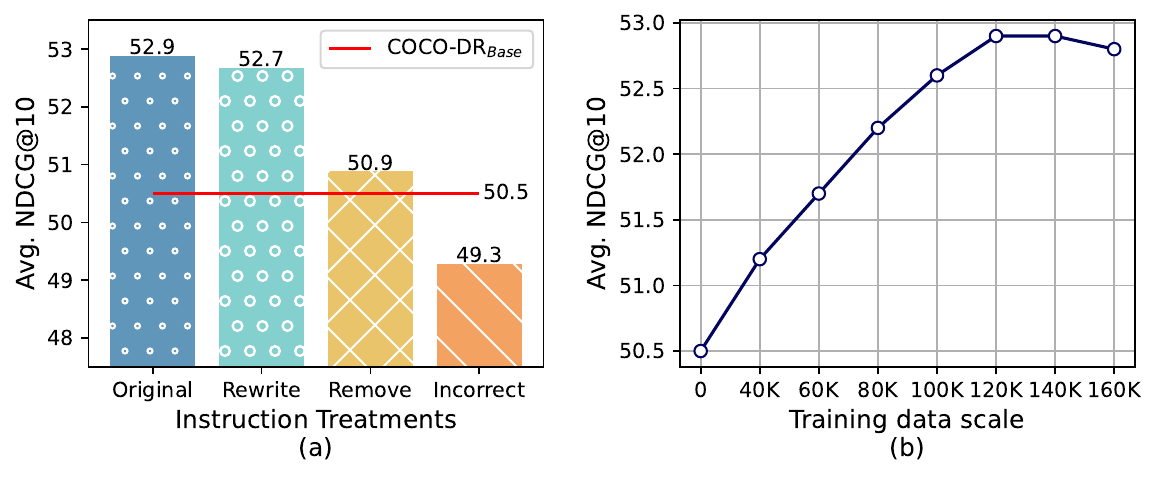}
    \vspace{-2.4em}
    \caption{(a): Performance on BEIR with different instruction treatments during testing. \textbf{(b):} Performance of \texttt{I3} with different training data scales.}\label{fig:abla2}
\vspace{-1.0em}
\end{figure}

The result of \textbf{Column 1} in Table~\ref{tab:ablation} indicates that directly instruction-tuning the backbone model actually undermines its inherent capabilities, resulting in a decline in performance. This observation underscores the crucial role of the \texttt{I3} architecture in enhancing zero-shot retrieval. 
The result of \textbf{Column 2} emphasizes the necessity of fine-tuning \texttt{I3} with instruction-based data. 
Besides, the result of \textbf{Column 3} confirms the superiority of iterative training with progressive structure pruning, which ensures that the knowledge learned by larger models is gradually and effectively transferred to smaller models.
Moreover, \textbf{Column 4} suggests that the process of data refinement is geared towards identifying and mitigating the drawbacks of the model, thus further boosting the performance. 
Lastly, the result of \textbf{Column 5} demonstrates that structure pruning effectively renders our model more lightweight, while simultaneously leading to almost no performance degradation compared to the un-pruned model.

\paragraph{\textbf{Retrieval Time Efficiency.}}
\textbf{Table~\ref{tab:effiency}} presents the average retrieval time efficiency across each dataset within BEIR, specifically comparing \texttt{I3}, COCO-DR, and the un-pruned version of \texttt{I3} (w/o pruning). 
The results reveal that the integration of the pluggable introspector, without pruning, significantly slows down the speed of query encoding by over 50\%. 
However, the incorporation of a pruned lightweight introspector only slightly decelerates the query encoding. 
Furthermore, the extra time incurred by the introspector has a minimal impact on the overall time, which is dominated by document encoding and the retrieval latency.
\begin{table}[t]
    \centering
    \caption{\label{tab:llama2}Performance on BEIR with different training data (ChatGPT-Generated, LLaMA2-Generated, from TART-dual).}
    \vspace{-1em}
    \begin{adjustbox}{width=0.42\textwidth}
    \begin{tabular}{l|cccc}
    \toprule
    Datasets & \tabincell{c} {Backbone\\COCO-DR} & \tabincell{c} {\texttt{I3}\\TART-Data}  & \tabincell{c} {\texttt{I3} \\LLaMA2-Data} & \tabincell{c} {\texttt{I3} \\ChatGPT-Data}\\
     \midrule
MS MARCO  &  {\bf{42.4}}  &  {---} &  {41.5} &  {41.8}  \\
     \midrule
TREC-COVID &  {80.4}  &  {78.2} &  {80.5} &  {\bf{81.6}}  \\
NFCorpus &  {35.4}  &  {36.0} &  {36.2} &  {\bf{37.1}} \\
NQ   &  {54.7}  &  {---} &  {55.9} &  {\bf{57.4}} \\
HotpotQA  &  {64.1}  &  {---} &  {\bf{64.3}} &  {63.3} \\
FiQA-2018   &  {32.9}  &  {34.6} &  {34.4} &  {\bf{35.7}}  \\
ArguAna   &  {51.5}  &  {56.1} &  {56.7} &  {\bf{59.8}}  \\
Touché-2020&  {26.3}  &  {23.9} &  {\bf{26.4}} &  {23.7}   \\
Quora&  {87.2}  &  {---} &  {87.5} &  {\bf{89.3}} \\
DBPedia-entity &  {40.7}  &  {41.1} &  {41.2} &  {\bf{41.8}} \\
SCIDOCS&  {17.8}  &  {18.7} &  {18.9} &  {\bf{19.9}} \\
Fever &  {79.3}  &  {---} &  {80.2} &  {\bf{80.8}}\\
Climate-Fever &  {30.4}  &  {29.6} &  {30.8} &  {\bf{31.1}} \\
SciFact   &  {72.2}  &  {76.7} &  {77.8} &  {\bf{79.9}} \\
CQADupStack  &  {39.3}  &  {---} &  {39.5} &  {\bf{40.0}}  \\    \midrule
\bf Avg TART Sub &  {43.1}  &  {43.9} &  {44.8} &  {\bf{45.6}} \\
\bf Avg &  {50.5}  &  {---} &  {52.2} &  {\bf{52.9}} \\
    \bottomrule
    \end{tabular}
    \end{adjustbox}
    
    \vspace{-1.3em}
\end{table}
\paragraph{\textbf{Incorporating with Different Backbones.} } 
As a generic approach, \texttt{I3} can be seamlessly integrated into different retrieval models with dual-tower architecture. Besides COCO-DR$_{\text{Large}}$, we also leverage COCO-DR$_{\text{Base}}$ and Contriever as the backbone to train \texttt{I3} in the same way. As shown in \textbf{Figure~\ref{fig:abla1}}, across all three backbone models, our method significantly enhances the zero-shot results, enabling them to achieve stronger performance. Notably, the performance of COCO-DR$_{\text{Base}}$, after instruction tuning within our framework, even surpasses the original COCO-DR$_{\text{Large}}$. This underscores the universality of our proposed approach.

\paragraph{\textbf{Impact of Instructions.}} To analyze the effectiveness of instructions, we employ different treatments for instructions during downstream zero-shot evaluation, \textit{i.e.}, \textbf{Rewrite Instr}, \textbf{Remove Instr}, and \textbf{Incorrect Instr}, which involve rephrasing the instructions without altering their original intents, entirely omitting the instructions, and providing misleading incorrect instructions, respectively. 
As shown in \textbf{Figure~\ref{fig:abla2}.a}, rewriting test instructions without altering their original meaning has minimal impact on performance. This demonstrates that \textit{our model is robust to various instructions}. 
However, the performance noticeably drops when no instructions are provided during evaluation, and providing instructions with incorrect retrieval intents leads to a more substantial decline in performance. 
This highlights \textit{the pivotal importance of instructions in intent-introspective retrieval}.

\paragraph{\textbf{Analysis of Training Data Sources.}}
Besides generating training data with ChatGPT, we also leverage \textit{\textbf{LLaMA2}} to synthesize an equivalent amount of training data. 
Additionally, we also experiment with \textit{\textbf{Berri}}, the manually-collected data in Tart-dual, as the training data to tune \texttt{I3} (BEIR datasets contained in \textit{Berri} are excluded in zero-shot evaluation). 
As shown in \textbf{Table~\ref{tab:llama2}}, replacing ChatGPT with LLaMA results in a marginal decrease in performance compared to the original \texttt{I3} due to reduced data quality, but it still surpasses COCO-DR and the model trained on Berri. The results show that our method does not simply rely on ChatGPT and highlight the efficacy of our data generation strategy.

\paragraph{\textbf{Impact of Training Data Scales}}
To investigate the impact of the training data scale, we control the total amount of training data by altering the size of the seed training dataset. 
As shown in \textbf{Figure~\ref{fig:abla2}.b}, the performance keeps increasing when the total number of training data expands from 40K to 120K.  
While further increasing the data size from 120K to 160K shows minimal impact on performance. 
These observations underscore that our iterative training paradigm is data-efficient.

\section{Conclusion and Future Work}

In this paper, we present \texttt{I3}, a generic and efficient approach capable of controlling retrieval models to introspect for specific retrieval intent and directly perform varied retrieval tasks without task-specific tuning. 
Integrating a pluggable introspector in a parameter-isolated manner, \texttt{I3} effectively preserves the original capability of the retrieval model, meanwhile efficiently empowering it with a new facet of intent-introspective retrieval conditioned on instructions. 
Furthermore,  we also innovatively propose progressively-pruned intent learning, which incorporates progressive structure pruning and drawback extrapolation-based data refinement, training \texttt{I3} phase-by-phase with extensive LLM-generated instruction data.
Extensive experiments demonstrate the superior zero-shot generalizability of \texttt{I3} on diverse retrieval tasks under both retrieval and reranking scenarios.

In the future, we aim to  extend the idea of instruction-based task-intent introspection across various fields to enhance the capabilities of different models~\cite{zhu2023graphcontrol, li2023fine, pan2023self, zhu2023sgl, li2020unsupervised}. 
Furthermore, we hope the technology of  intent-introspective retrieval technology can emerge as  an important tool for augmenting a range of  downstream tasks, such as LLM pre-training~\cite{zhang2024recost, li2022fine}, video comprehension~\cite{li2021adaptive,li2022compositional,li2023variational, li2022dilated}, and anomaly detection~\cite{chen2023improving, guo2023rustgraph}.

\section*{Acknowledgements}
This work was supported by the NSFC (No. 62272411), Key Research and Development Projects in Zhejiang Province (No. 2024C01106), the National Key Research and Development Project of China (2018AAA0101900), Alibaba-Zhejiang University Joint Research Institute of Frontier Technologies, and Ant Group.


\bibliographystyle{ACM-Reference-Format}
\bibliography{sample-base}

\newpage
\appendix
\section*{Appendix}

\definecolor{mycolor}{rgb}{0.122, 0.435, 0.698}
\newtcolorbox{mybox}{colframe =mycolor}

\section{Detailed Prompt for Data Synthesizing}
\begin{figure}[t]
    \includegraphics[width=\linewidth]{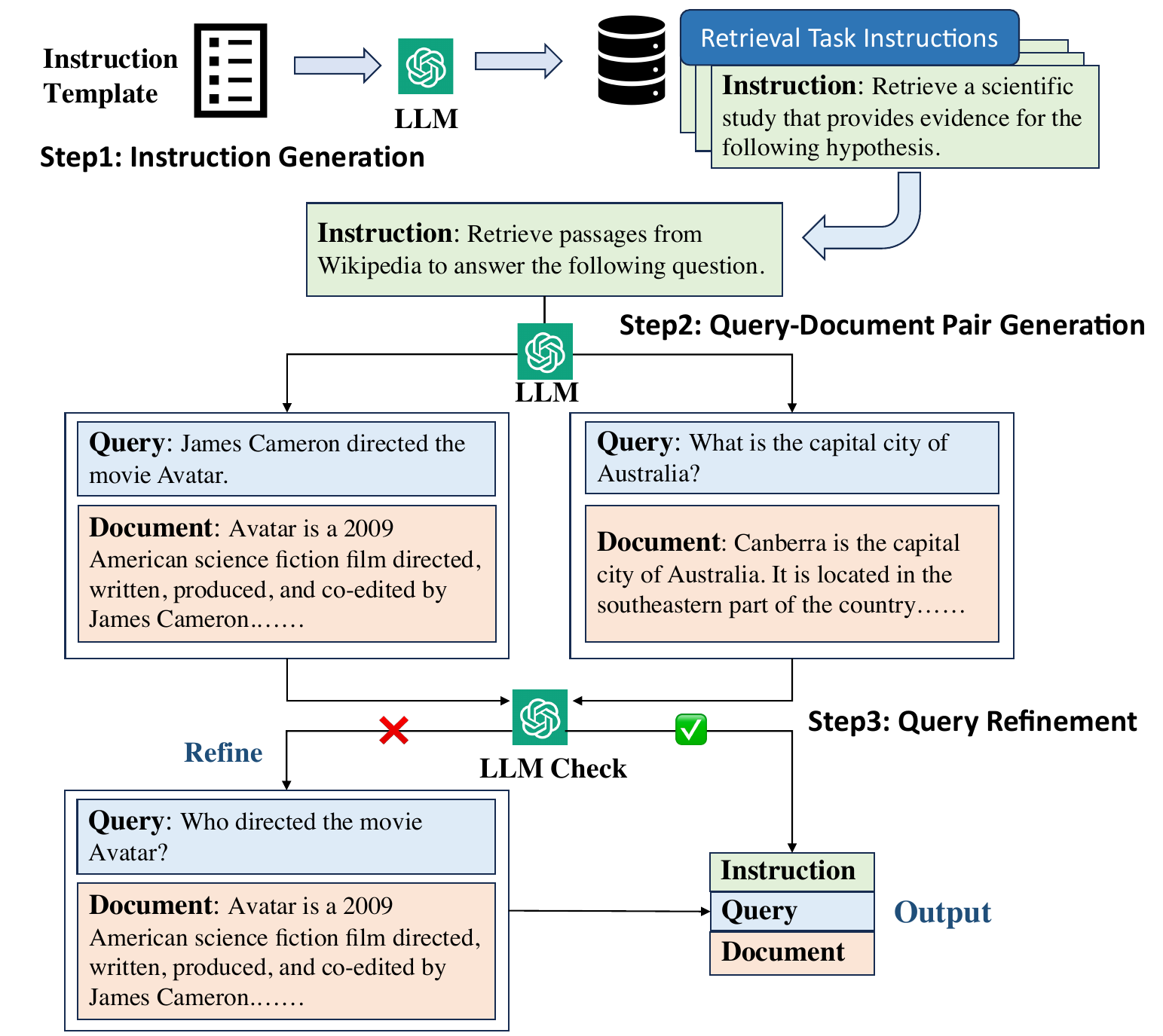}
    \caption{LLM-guided instruction data synthesizing.}\label{fig:synthesizing}
\vspace{-1.3em}
\end{figure}

The LLM-guided instruction data synthesizing involves three steps as shown in Figure \ref{fig:synthesizing}: 1) instruction generation; 2) query-document pair generation; 3) query self-check (refinement).

\paragraph{\textbf{Instruction Generation.}}We first utilize the following prompt template to generate diverse instructions covering a wide range of retrieval intents:
\begin{mybox}
The task is to generate some diverse instructions that reveal search intents for retrieval tasks. Here are some generation requirements:\newline
(1) Each generated instruction must explicitly outline the retrieval intent, which describes how the retrieved text relates to the query, such as whether the text answers a question in the query. \newline
(2) Within each generated instruction, you must specify the expected source or topic of retrieved text, such as Wikipedia, scientific, or legal. \newline
(3) Within each generated instruction, you should also define the text block to retrieve, such as a document or a paragraph.\newline
Here are specific examples of instructions: \newline
$<$\texttt{GIVE INSTRUCTION EXAMPLES HERE}$>$ \newline
Now please directly generate instructions without writing any other explanations:
\end{mybox}

\paragraph{\textbf{Query-Document Pair Generation. }} For each generated instruction, we first extract the \texttt{TEXT TYPE} of the retrieved text, including the topic (\textit{e.g.}, scientific, legal) and the organizational formats (\textit{e.g.}, document, paragraph). Subsequently, we employ the following prompt template to generate query-document pairs for each instruction:
\begin{mybox}
I will give you an instruction that describes how the retrieved text relates to the query in a retrieval task. The task is to generate a pair of query and the retrieved text based on the given instruction. Here are some generation requirements: \newline
(1) The retrieved text you generate should be $<$\texttt{TEXT TYPE}$>$ according to the instruction.\newline
(2) The connection between your generated query and the retrieved text should correspond to the relationship specified in the instruction.\newline
Here are examples of generating query and retrieved text with instructions: \newline
$<$\texttt{EXAMPLES OF QUERY-DOCUMENT PAIR GENERATION}$>$ \newline 
\newline
Please directly generate the query and the retrieved text without writing any other explanations, based on the following instructions:\newline
$<$\texttt{INSTRUCTION}$>$
\end{mybox}

\paragraph{\textbf{Query Self-check. }}We observed that for certain generated instructions, some of the subsequent generated queries and documents do not always correctly capture the retrieval intents expressed within those instructions. To address this challenge, we finally design the following prompt templates to refine or regenerate these misaligned queries:
\begin{mybox}
 I will give you an instruction that describes how the retrieved text relates to the query in a retrieval task. And then you are provided with the query and the retrieved text. You should assess the query for the following criteria: \newline
 (1) Does the given query adhere to the query format specified in the instruction? \newline
(2) Does the relationship between the query and the retrieved text align with the relationship specified in the instruction? \newline
If the existing query satisfies the aforementioned criteria, directly output the existing query without any alterations. Conversely, if the query falls short of meeting the criteria, please revise or rewrite the existing query to make it satisfy the criteria and then directly output the revised query. DO NOT provide any explanations. \newline
Here is an example: \newline
$<$\texttt{EXAMPLES OF QUERY REFINEMENT}$>$ \newline \newline
Now I will give you the instruction, the retrieved text, and the query: \newline
The instruction: $<$\texttt{INSTRUCTION}$>$ \newline
The retrieved text: $<$\texttt{RETRIEVED TEXT}$>$ \newline
The query: $<$\texttt{QUERY}$>$
\end{mybox}

\section{Evaluation Instructions}
\label{app:bbd}

\begin{table}[t!]
\centering
\caption{\label{tab:test_instruction}Evaluation instructions for BEIR benchmark.}
\vspace{-1em}
\footnotesize
\begin{tabular}{lp{6cm}}
\toprule
\textbf{Dataset} & \textbf{Instruction} \\\midrule
MS MARCO & \texttt{I want to know the answer to the question. Can you find good evidence on the web?} \\ \midrule
TREC-COVID & \texttt{Retrieve Scientific paper paragraph to answer this question.} \\ \midrule
NFCorpus & \texttt{Retrieve Scientific paper paragraph to answer this question.} \\ \midrule
NQ & \texttt{retrieve passages from Wikipedia that provides answers to the following question.}\\ \midrule
HotpotQA & \texttt{Find a paragraph that provides useful information to answer this question.}\\ \midrule
FiQA-2018 & \texttt{Find financial web article paragraph to answer.}\\ \midrule
ArguAna & \texttt{Retrieve an argument that counter argues the following paragraph.}\\ \midrule
Touché-2020 & \texttt{You have to retrieve an argument to this debate question.} \\ \midrule
Quora & \texttt{Check if a Quora question is duplicated with this question.}\\ \midrule
DBPedia-entity & \texttt{Retrieve a Wikipedia introduction paragraph of the following entity.}\\ \midrule
SCIDOCS & \texttt{Find scientific paper titles that are related to the following.}\\ \midrule
Fever & \texttt{Retrieve a Wikipedia paragraph to verify this claim.}\\ \midrule
Climate-Fever & \texttt{Retrieve a Wikipedia paragraph to verify this claim.}\\ \midrule
SciFact & \texttt{Retrieve a scientific paper sentence to verify if the following claim is true.}\\ \midrule
CQADupStack & \texttt{I want to identify duplicate questions asked in community question answering forums.}\\
\bottomrule
\end{tabular}
\vspace{-1.6em}
\end{table}

We conduct all experiments on the BEIR, a widely used benchmark for zero-shot evaluation of information retrieval models. We follow the evaluation instructions in \citet{tart} to provide an extra instruction for each dataset to reflect specific retrieval intents. The detailed evaluation instructions can be found in Table~\ref{tab:test_instruction}.

\end{document}